\documentclass[letterpaper, 10 pt, conference]{ieeeconf}  

\IEEEoverridecommandlockouts                              

\usepackage[printwatermark]{xwatermark}
\usepackage{xcolor}
\usepackage{multirow}
\usepackage{amsmath}
\usepackage{amsthm}
\usepackage{amssymb}
\usepackage{graphicx} 
\usepackage{subfig}
\usepackage{booktabs}
\usepackage[ruled]{algorithm2e}
\usepackage{flushend}
\usepackage[font=small,skip=0pt]{caption}
\usepackage{cite}
\newtheorem{theorem}{Theorem}

\theoremstyle{definition}

\newcommand{\mb}[1]{\mathbf{ #1 }}

\usepackage{array}
\newcolumntype{C}[1]{>{\centering\arraybackslash}p{#1}}
\begin{document}

\title{\LARGE \bf
Koopman NMPC: Koopman-based Learning and Nonlinear Model Predictive Control of Control-affine Systems
}

\author{Carl Folkestad$^*$ and Joel W. Burdick$^*$\thanks{$^*$Both authors are with the Division of Engineering and Applied Sciences, California Institute of Technology, Pasadena, CA, USA}}

\maketitle
\thispagestyle{empty}
\pagestyle{empty}

\begin{abstract}
Koopman-based learning methods can potentially be practical and powerful tools for dynamical robotic systems. However, common methods to construct Koopman representations seek to learn \textit{lifted linear} models that cannot capture nonlinear actuation effects inherent in many robotic systems. This paper presents a learning and control methodology that is a first step towards overcoming this limitation. Using the Koopman canonical transform, control-affine dynamics can be expressed by a {\em lifted} \textit{bilinear} model. The learned model is used for nonlinear model predictive control (NMPC) design where the bilinear structure can be exploited to improve computational efficiency. The benefits for control-affine dynamics compared to existing Koopman-based methods are highlighted through an example of a simulated planar quadrotor. Prediction error is greatly reduced and closed loop performance similar to NMPC with full model knowledge is achieved. 
\end{abstract}

\section{Introduction}
\label{sec:introduction}

Efficient control design for dynamic robotic systems is a persistent challenge when optimal task performance is sought while satisfying state, actuation, and computing constraints. Because of its ability to intuitively specify both performance metrics and constraints, model predictive control (MPC) is attractive. Advances in optimization algorithms and computing power are enabling \textit{nonlinear} MPC (NMPC) to be deployed on robotic systems in real-time if carefully implemented \cite{Kouzoupis2018, Gros2020, Grandia2020}. One key challenge in developing NMPC, however, is to obtain a sufficiently accurate model of the system to be controlled. {\em Learning} algorithms can capture the salient aspects of a robot's complex mechanics and environmental interactions, thereby reducing the need for laborious system identification procedures \cite{Lupashin2014}. Many approaches have been proposed to learn models intended for control design (cf. \cite{GaussianDynamic, Shi2018, Taylor2019a, Recht2019, Kaiser}). However, using these models for NMPC is not straightforward as they are either unsuitable or expensive to simulate and/or discretize to solve the resulting nonlinear program. We take a Koopman-centric approach to learn a {\em lifted bilinear model} of the dynamics that can be incorporated in NMPC to design close to optimal controllers that incorporate state and actuation constraints.

We are interested in learning control-affine dynamics, $\dot{x} = f(x) + g(x)u$, which allow a wide class of aerial and ground robots to be characterized. Conventionally, a control system's behavior is studied via its state space flows. In contrast, Koopman approaches study the evolution of {\em observables}, which are functions over the state-space. In this space, an autonomous dynamical system can be represented by a {\em linear} (but possibly infinite dimensional) operator \cite{Lan2013, mauroy2016linear}. Data-driven methods for identifying Koopman models have received considerable attention. Dynamic Mode Decomposition (DMD) and extended DMD (EDMD) have been shown to efficiently identify finite dimensional approximations of the Koopman operator of the system dynamics \cite{Schmid2010, Williams2015, Korda2018a, Taira2017, Berger2015, Brunton2016}. 

Koopman-style modeling has also been extended to {\em controlled} nonlinear systems \cite{Kaiser, Proctor2018}. However, most current methods approach this problem by learning a lifted \textit{linear} model, which only allows the control vector fields, $g(x)$, to take a constant  state-invariant form.  This is a significant limitation, as many useful robotic systems, e.g. systems where input forces enter the system dynamics through rotation matrices, are best described by nonlinear control-affine dynamics. To overcome this limitation, we learn a model motivated by the \textit{Koopman canonical transform} (KCT) \cite{Surana2016}, which allows a large class of nonlinear control-affine dynamic models to be \textit{lifted} to a higher-dimensional space where the system evolution can be described by a \textit{bilinear} (but possibly infinite dimensional) dynamical system. This lifting is achieved by careful design of the function dictionary and employing an EDMD learning method \cite{Williams2015, Korda2018a}.

Prior work on Koopman-based control design has primarily focused on applying linear MPC to lifted linear models, and has been successfully implemented in both simulated and robotic experiments \cite{Korda2018a, Folkestada, Bruder2019}. Design with bilinear models is less explored, but connections between Koopman bilinear system descriptions and classical control concepts such as reachability and control Lyapunov functions have been presented \cite{Goswami2018, Huang2019}. Very recently, bilinear Koopman models linearized at the current state of the system were used in MPC \cite{Bruder2020}. Another approach uses the bilinear model structure to simplify the construction of a control Lyapunov function enforced as a constraint in a nonlinear MPC method to obtain stability guarantees \cite{Narasingam2020}.

While a few works have addressed NMPC design for bilinear Koopman models \cite{Bruder2020}, \cite{Narasingam2020}, little consideration has been given to practical real-time realization of these methods on robotic systems, which often require high control rates due to fast dynamics. Towards this goal, this paper presents Koopman NMPC, combining the process of learning control-affine dynamics in Koopman bilinear form with NMPC design. Our contributions are twofold. First, building on recent advances in NMPC, we develop a controller for bilinear Koopman models that uses the bilinear model structure to improve computational efficiency, making real-time computation possible. Second, we show the advantages of learning lifted bilinear models over linear models and demonstrate that the completely data-driven Koopman NMPC method can  match the performance of a NMPC controller with full a priori model knowledge on a simulated planar quadrotor.  

This manuscript is organized as follows. Preliminaries on the KCT and NMPC is presented in Section \ref{sec:preliminaries}. Then, the learning method to learn lifted bilinear models is described in Section \ref{sec:bedmd}, and the Koopman NMPC developed in Section \ref{sec:mpc}. Finally, the method is demonstrated on a simulated planar quadrotor in Section \ref{sec:result} before we conclude in Section \ref{sec:conclusion}.

\section{Preliminaries}
\label{sec:preliminaries}

We consider control-affine continuous-time dynamical systems of the form 
\begin{equation}\label{eq:dynamics}
        {\dot{x}} = f(x) + \sum_{i=1}^m g_i(x)u_i
\end{equation}
\noindent where $x \in \mathcal{X} \subseteq \mathbb{R}^d, u \in \mathcal{U} \subseteq \mathbb{R}^m$, and $f, g_i, i=1,\dots,m$ are assumed to be Lipschitz continuous on $\mathcal{X}, \mathcal{U}$.

\subsection{Koopman spectral theory}
Before considering the effects of control inputs, we introduce the Koopman operator, which is defined for autonomous continuous-time dynamical systems:
\begin{equation}\label{eq:aut_dynamics}
    \dot{x} = f_{aut}(x)
\end{equation}
with state $x \in \mathcal{X} \subset \mathbb{R}^d$, and $f$ is assumed to be Lipschitz continuous on $\mathcal{X}$. The flow of  (\ref{eq:aut_dynamics}) is denoted by $S_t(x)$, defined as $\frac{d}{dt}S_t(x) = f_{aut}(S_t(x))$ for all $x \in \mathcal{X}, t\geq 0$. The \textit{Koopman operator semi-group} $(U_t)_{t\geq0}$, from now on simply denoted as the \textit{Koopman operator}, is defined as 
\begin{equation}
    U_t \varphi = \varphi \circ S_t
\end{equation}
for all $\varphi \in \mathcal{C}(\mathcal{X})$, where $\mathcal{C}(\mathcal{X})$ is the space of continuous observables $\varphi: \mathcal{X} \rightarrow \mathbb{C}$, and $\circ$ denotes function composition. Each element of $U_t: \mathcal{C}(\mathcal{X}) \rightarrow \mathcal{C}(\mathcal{X})$ is a \textit{linear} operator. 

An \textit{eigenfunction} of the Koopman operator associated to an eigenvalue $\lambda \in \mathbb{C}$ is any function $\phi \in \mathcal{C}(\mathcal{X})$ that defines a coordinate evolving linearly along the flow of (\ref{eq:aut_dynamics})
\begin{equation}
    (U_t \phi)(x) = \phi(S_t(x)) = e^{\lambda t} \phi (x).
\end{equation}



\subsection{The Koopman canonical transform}
We now return to control-affine dynamics (\ref{eq:dynamics}) and recall when and how they can be transformed to a \textit{bilinear} form through the \textit{Koopman canonical transform} \cite{Surana2016}. Let $\big(\lambda_i, \phi_i(x) \big), i = 1,\dots,n$ be eigenvalue-eigenfunction pairs of the Koopman operator associated with the autonomous dynamics of (\ref{eq:dynamics}), $\dot{x} = f(x)$. The KCT relies on the assumption that the state vector can be described by a finite number of eigenfunctions, i.e. that $ x = \sum_{i=1}^n \phi_i(x) v_i^{x}$ for all $ x\in \mathcal{X}$, and where $v_i^{x} \in \mathbb{C}^d$. This is likely to hold if $n$ is large. If not, they may be well approximated by $n$ eigenfunctions.

When $\phi_i: \mathcal{X} \rightarrow \mathbb{R}$, the KCT is defined as
\begin{align}\label{eq:kct}
        x = C^{x} z, \quad \dot{z} = Fz + \sum_{i=1}^mL_{g_i}T(x)u_i
\end{align}
\noindent where $z = T(x) = [\phi_1(x) \dots \phi_n(x)]^T$, $C^{x} = [v_1^{x}\dots v_n^{x}]$, and $F \in \mathbb{R}^{n \times n}$ is a diagonal matrix with entries $F_{i,i} = \lambda_i$. 

Under certain conditions the system (\ref{eq:kct}) is bilinearizable in a countable, possibly infinite basis. We restate the conditions for the existence of a bilinear form in a \textit{finite} basis as this is of practical interest in the following theorem. 

\begin{theorem} \label{th:kbf}
\cite{Goswami2018} Suppose there exist Koopman eigenfunctions $\phi_j, j=1,\dots,n, n \in \mathbb{N}, n < \infty$ of the autonomous dynamics (\ref{eq:dynamics}) whose span, $span(\phi_1,\dots,\phi_n)$, forms an invariant subspace of $L_{g_i}, i = 1,\dots,m$. Then, the system (\ref{eq:dynamics}), and in turn system (\ref{eq:kct}), are bilinearizable with an n-dimensional state space.
\end{theorem}

Although the conditions of Theorem \ref{th:kbf} may be hard to satisfy in a given problem, an approximation of the true system (\ref{eq:dynamics}) can be obtained with sufficiently small approximation error by including adequately many eigenfunctions in the basis. As a result, $L_{g_i} = G_i$ and the system can be expressed as the \textit{Koopman bilinear form} (KBF) (see \cite{Goswami2018} for details): 
\begin{equation} \label{eq:zdot}
    \dot{z}=F z+\sum_{i=1}^m G_i z u_i, \quad z \in \mathbb{R}^n, n < \infty .
\end{equation}

\subsection{Nonlinear model predictive control}\label{sec:nmpc}
When the exact continuous dynamics (\ref{eq:dynamics}) are known, the general optimal control problem is intractable because there are infinitely many optimization variables. To reformulate the problem into a tractable finite-dimensional nonlinear program (NLP) we discretize the dynamics. Given a time horizon $T$, consider the time increment $\Delta t$ and divide the time horizon [0,T] into $N = \frac{T}{{\Delta t}}+1$ discrete subintervals $[t_k,t_{k+1}], t_k = k \Delta t, k\!=\!0,\!\dots\!,N\!-\!1$. Replacing the continuous control signal $u(t)$ with a zero-order-hold signal, the dynamics are integrated over each interval with an appropriate integration scheme to get a discrete-time representation of the dynamics $x_{k+1} = f_d(x_k, u_k)$, where $x_k = x(t_k)$. 

The quadratic objective NMPC problem is formulated as:
\begin{align} \label{eq:nonlin_mpc}
    \begin{split}
        \min_{\mb x, \mb u} \quad & \sum_{k=0}^{N-1} \frac{1}{2} \begin{bmatrix}
        x_k - x_k^\text{ref}\\ u_k - u_k^\text{ref} 
        \end{bmatrix}^T W_k \begin{bmatrix}
        x_k - x_k^\text{ref}\\ u_k - u_k^\text{ref} 
        \end{bmatrix} \\
        \text{s.t.} \qquad & x_{k+1} = f_d(x_k, u_k), \quad k=0,\dots,N-1\\
        & x_0 = \hat{x}, \quad h_k(x_k, u_k) \leq 0, \quad \,\, k = 0,\dots,N\\
    \end{split}
\end{align}
where $\hat{x}$ is the current state, $\mb x = [x_0, \dots, x_N], \mb u = [u_0, \dots, u_{N-1}]$ are the stacked matrices of state and control vectors for each time step, $W$ is the positive semi-definite cost matrix, and $h_k$ is the constraint function encoding state and actuation constraints, both allowed to change at every timestep. In classical receding horizon fashion, at each timestep, a new state estimate $\hat{x}$ is obtained, the optimization problem is solved, and the control signal solution corresponding to the first timestep, $u_0$, is deployed to the system.

\subsection{Sequential quadratic programming}\label{sec:sqp}
NMPC problems (\ref{eq:nonlin_mpc}) are primarily solved via interior point (IP) or sequential quadratic programming (SQP) methods. SQP-approaches can leverage the fact that the nonlinear programming problems (NLP) solved at adjacent timesteps are quite similar, so that the solution of the NMPC problem at the previous timestep can be used to \textit{warm-start} the solution at the current timestep. This warm-start feature greatly reduces the real-time computational burden, and often a single SQP iteration is sufficient at each timestep to arrive at a close-to-optimal solution of the NMPC problem \cite{Kouzoupis2018}. 

\begin{algorithm}[b]
\small
\SetAlgoLined
\textbf{Input:} current state $\hat{x}_i$, reference trajectory $(\mb x_i^{\text{ref}}, \mb u_i^{\text{ref}})$, initial guess $(\mb x_i^{\text{init}}, \mb u_i^{\text{init}})$\\
\While{Not converged}{
Form $r\!_{i,k}, h_{i,k}, A_{i,k}, B\!_{i,k}, C\!_{i,k}, D\!_{i,k}, H\!_{i,k}, J\!_{i,k}$ by (\ref{eq:linearization})\\
Solve (\ref{eq:qp-mpc}) to get the Newton direction $(\Delta \mb x_i, \Delta \mb u_i)$\\
Update initial guess with the Newton step:
$(\mb x_i^{\text{init}}, \mb u_i^{\text{init}}) \leftarrow (\mb x_i^{\text{init}} + \Delta \mb x_i, \mb u_i^{\text{init}}+ \Delta \mb u_i)$
}
\textbf{Return:} NMPC solution $(\mb x_i, \mb u_i) = (\mb x_i^{\text{init}}, \mb u_i^{\text{init}})$
\caption{\cite{Gros2020} SQP for NMPC at discrete time $i$}
\label{algo:sqp}
\end{algorithm}

In the SQP algorithm, summarized in Algorithm \ref{algo:sqp}, Eq. (\ref{eq:nonlin_mpc}) is sequentially approximated by quadratic programs (QPs), whose solutions are Newton directions for performing steps toward the optimal solution of the NLP. The sequence is initialized at an initial guess of the solution, $(\mb x_0^{\text{init}}, \mb u_0^{\text{init}})$, at which the following QP is iteratively solved and the initial guess updated at each iteration \textit{i} until convergence:
\begin{align} \label{eq:qp-mpc}
    \begin{split}
        &\min_{\Delta \mb x_i, \Delta \mb u_i} \quad  \sum_{k=0}^N \begin{bmatrix} \Delta \mb x_{i,k}\\ \Delta \mb u_{i,k} \end{bmatrix}^T H_{i,k} \begin{bmatrix} \Delta \mb x_{i,k}\\ \Delta \mb u_{i,k} \end{bmatrix} + J_{i,k}^T \begin{bmatrix} \Delta \mb x_{i,k}\\ \Delta \mb u_{i,k} \end{bmatrix}\\
        \text{s.t.} \,& \Delta \mb x_{i,k+1} \!=\! A_{i,k} \Delta \mb x_{i,k} \!+\! B_{i,k}\Delta \mb u_{i,k} + r_{i,k}, \, k\!=\!0,...,N\!\!-\!\!1,\\
        & C_{i,k} \Delta \mb x_{i,k} + D_{i,k} \Delta \mb u_{i,k} + h_{i,j} \leq 0, \qquad \,\, k = 0,...,N,\\
        & \Delta \mb x_{i,0} = \hat{x}_i - \mb x_{i,0}^{\text{init}},
    \end{split}
\end{align}
where $H_{i,k}$ is the Hessian of the NLP Lagrangian (\ref{eq:nonlin_mpc}) and
\begin{align} \label{eq:linearization}
    \begin{split}
        A_{\!i\!,k}\! &=\! \frac{\partial f_{\!d}}{\partial x} \! \bigg |_{\!\begin{smallmatrix}\mb x_i^{\text{init}}\\ \mb u_i^{\text{init}}
        \end{smallmatrix}}\!, \,
        B_{\!i\!,k}\! =\! \frac{\partial f_{\!d}}{\partial u} \! \bigg |_{\!\begin{smallmatrix}\mb x_i^{\text{init}}\\ \mb u_i^{\text{init}}
        \end{smallmatrix}}\!, \,
        C_{\!i\!,k}\! =\! \frac{\partial h}{\partial x} \! \bigg |_{\!\begin{smallmatrix}\mb x_i^{\text{init}}\\ \mb u_i^{\text{init}}
        \end{smallmatrix}}\!, \, 
        D_{\!i\!,k}\! =\! \frac{\partial h}{\partial u} \! \bigg |_{\!\begin{smallmatrix}\mb x_i^{\text{init}}\\ \mb u_i^{\text{init}}
        \end{smallmatrix}}\!,\\
        r_{i,k} &= f_d(\mb x_{i,k}^{\text{init}},\mb u_{i,k}^{\text{init}}) - \mb x_{i,k+1}^{\text{init}}, \quad
        h_{i,k} = h(\mb x_{i,k}^{\text{init}},\mb u_{i,k}^{\text{init}}),\\
        J_{i,k} &= W_{i,k}\begin{bmatrix} \mb x_{i,k}^{\text{init}} - \mb x_{i,k}^{\text{ref}}\\
        \mb u_{i,k}^{\text{init}} - \mb u_{i,k}^{\text{ref}}\end{bmatrix}.
    \end{split}
\end{align}


\section{Learning Lifted Bilinear Dynamics}
\label{sec:bedmd}

\subsection{Modeling assumptions and data collection}

We use EDMD to learn approximate finite dimensional lifted bilinear dynamics from data. The system's unknown dynamics are assumed to be control-affine, with $f, g_1, \dots, g_m$ in (\ref{eq:dynamics}) unknown. We seek to learn a model and design a multi-purpose controller for the unknown system.

We assume that a nominal controller permits us to execute  $M_t$ data collection trajectories of length $T_t$ from initial conditions $x_0^j \in \Omega, j=1,\dots,M_t$. From each trajectory, $M_s = (T_t/\Delta t)$ state and control actions are sampled at a fixed time interval $\Delta t$, resulting in a data set
\begin{equation}
    \mathcal{D} = \bigg ( \big (x_k^j, u_k^j \big )_{k=0}^{M_s} \bigg ) _{j=1}^{M_t}.
\end{equation}

Since the NMPC design requires continuous-time models to be discretized, we learn a discrete-time lifted bilinear model, thereby avoiding potential numerical differentiation and discretization errors. This is further motivated by the existence of discretization procedures that maintain stability properties and the bilinear structure of the original system, such as the trapezoidal rule with zero-order-hold \cite{Phan2012b, Surana2018}. 


\subsection{Supervised learning of unknown dynamics}

Define a dictionary of $D$ dictionary functions $z = \phi(x), \, \phi: \mathbb{R}^d \rightarrow \mathbb{R}^D$. The choice of the functions can be based on system knowledge (i.e. feature engineering) or be a generic basis of functions such as monomials of the state up to a certain degree. Choosing dictionary functions is an ongoing area of research in Koopman-based learning methods. The most promising and principled choice of functions is arguably using data-driven Koopman eigenfunctions \cite{ACC2019, Korda2019}. However, existing methods cannot be readily applied in the KCT-setting and extending the methods to control-affine dynamics is outside of the scope of this paper.

To learn a lifted bilinear dynamic model, the data $\mathcal{D}$ is organized into data matrices $X, X', U$, where each corresponding column of $X$, and $X'$ are state samples recorded one sampling interval apart, see (\ref{eq:bilinear_regression}). Then, the lifted data matrix is created by applying $\phi(x)$ to each column of $X$ and $X'$, denoted $Z = \phi(X), Z' = \phi(X')$ by slight abuse of notation. Finally, $Z_{u}$ is constructed by applying $\phi_{u}(x, u)$ to each corresponding pair of columns of $X$ and $U$, where $\phi_{u}(x, u) = [\phi(x) \,\, \phi(x)u_1 \, \dots \, \phi(x)u_m]^T$. Learning can then be formulated as a linear regression problem (\ref{eq:bilinear_regression}). 
\begin{align}\label{eq:bilinear_regression}
\begin{split}
    \min_{F, G_1,\dots, G_m \in \mathbb{R}^{D \times D}} &||Z' - \begin{bmatrix} F & G_1 & \dots & G_m \end{bmatrix} Z_{u} ||^2\\
\min_{C^{x} \in \mathbb{R}^{d \times D}} &||X - C^{x} Z ||^2
\end{split}
\end{align}
\vspace{-0.5cm}
\begin{align*}
    &X = \begin{bmatrix} x_0^1 & \dots  x_{M_s-1}^1 & \dots &  x_0^{M_t} & \dots  x_{M_s-1}^{M_t} \end{bmatrix},\\
    &X' = \begin{bmatrix}  x_1^1 & \dots  x_{M_s}^1 & \dots &  x_1^{M_t} & \dots  x_{M_s}^{M_t} \end{bmatrix},\\
    &U = \begin{bmatrix}  u_0^1 & \dots  u_{M_s-1}^1 & \dots &  u_0^{M_t} & \dots  u_{M_s-1}^{M_t} \end{bmatrix}, \\
    &Z = \phi(X), \, Z' = \phi(X'), \, Z_{ u} = \phi_{u}(X, U)\ .
\end{align*}
Regularization, such as sparsity-promoting $l_1$-regularization which has been shown to improve prediction performance and reduce overfitting \cite{Kaiser}, can be added to the regression. Furthermore, learning $C^{ x}$ is not needed if the projection from the lifted space to the original space can be analytically computed for the chosen dictionary. For example, a monomial basis will typically include the state itself. This results in a lifted discrete-time bilinear model of the form
\begin{align}
        x_k = C^{ x},  z_k \quad z_{k+1} = F z_k + \sum_{l=1}^m G_l  z_k u_{k,l}.
\end{align}


\section{Nonlinear model predictive control design}
\label{sec:mpc}

\subsection{Design considerations}
Based on Section \ref{sec:preliminaries}, we first reformulate the NMPC problem (\ref{eq:nonlin_mpc}) using the identified Koopman bilinear model: 
\begin{align} \label{eq:koopman_nmpc}
    \begin{split}
        \min_{Z, U} \quad & \sum_{i=0}^{N} \begin{bmatrix} \mb z_{i,k}^{\text{init}} +\Delta \mb z_{i,k}\\ \mb u_{i,k}^{\text{init}} +\Delta \mb u_{i,k} \end{bmatrix}^T W_{i,k} \begin{bmatrix} \mb z_{i,k}^{\text{init}} + \Delta \mb z_{i,k}\\ \mb u_{i,k}^{\text{init}} + \Delta \mb u_{i,k} \end{bmatrix}\\
        \text{s.t.} \qquad &  z_{k+1} = F  z_k + \sum_{i=1}^m G_i  z_k  u_k^{(i)}, \quad k=0,\dots,N-1\\
        & c_l \leq C^{ x}  z_k \leq c_u, \quad d_l \leq  u_k \leq d_u, \quad  k = 0,\dots,N,\\
        &  z_0 = \phi(\hat{ x}) .
    \end{split}
\end{align}
The initialization and closed loop operation of the controller can be summarized as follows (see Algorithm \ref{algo:koopman_mpc}). Before task execution, the SQP algorithm with the Koopman QP subproblem (\ref{eq:koopman_nmpc}) is executed to convergence to obtain a good initial guess of the solution. Then, in closed loop operation, the Koopman bilinear model is linearized along the initial guess, the current state is obtained from the system, the current state is lifted using the function basis, and then the QP subproblem is solved only once. Finally, the first control input of the optimal control sequence is deployed to the system, and the full solution is shifted one timestep and used as an initial guess at the next timestep.

\begin{algorithm}[t]
\small
\SetAlgoLined
\textbf{Input:} reference trajectory $(\mb x_i^{\text{ref}}\!\!\!, \mb u_i^{\text{ref}})$, initial guess $(\mb x_i^{\text{init}}\!\!\!, \mb u_i^{\text{init}})$\\
\While{Controller is running}{
Form $r_{i,k}, A_{i,k}, B_{i,k}$ using (\ref{eq:koopman_linearization})\\
Get and lift current state, $z_{i,0} = \phi(\hat{x})$\\
Solve (\ref{eq:koopman_nmpc}) to get the Newton direction $(\Delta \mb x_i, \Delta \mb u_i)$\\
Update solution, 
$(\mb x_i, \mb u_i)\! \leftarrow\! (\mb x_i^{\text{init}} \!+\! \Delta \mb x_i, \mb u_i^{\text{init}}\!+\! \Delta \mb u_i)$\\
Deploy first input $u_0$ to the system\\
Construct $(\mb x_{i+1}^{\text{init}}, \mb u_{i+1}^{\text{init}})$ using (\ref{eq:shift_procedure})
}
\caption{Koopman NMPC (closed loop)}
\label{algo:koopman_mpc}
\end{algorithm}

 
Although we have restricted the objective to be quadratic and the state and actuation constraints to be linear (except for the evolution of the dynamics), nonlinear objective and constraint terms can be included by adding them to the lifted state $ z = \phi( x)$. For example, if it is desired to enforce the constraint $\cos(x_1) \leq 0$, we can add $\phi_j = \cos(x_1)$ to the lifted state and enforce $ z_k^{(j)} \leq 0$ \cite{Korda2018a}.

While not our main focus, a discussion of how to achieve guaranteed closed loop stability of the proposed control strategy is in order. In the nominal case, with no model mismatch between the true dynamics and the Koopman bilinear model, closed loop stability of the controller for bilinear systems with a quasi-infinite method has been shown (see e.g. \cite{Bloemen2002}). 
More recently, some early stability results using Lyapunov MPC methods have been developed \cite{Narasingam2020}. In particular, the bilinear model structure simplifies the construction of a Lyapunov function that is added as a constraint to the MPC. Lyapunov stability of the controller based on the KBF is then proved under the assumption that the prediction error of the learned Koopman model is finite. Although promising, further analysis of robustness and stability properties of the methodology is needed. In this work however, we focus on the practical implementation and defer further theoretical development to future research.

\subsection{Warm-start of SQP at each timestep}

\begin{figure*}[t]
    \centering
    \includegraphics[width=\textwidth]{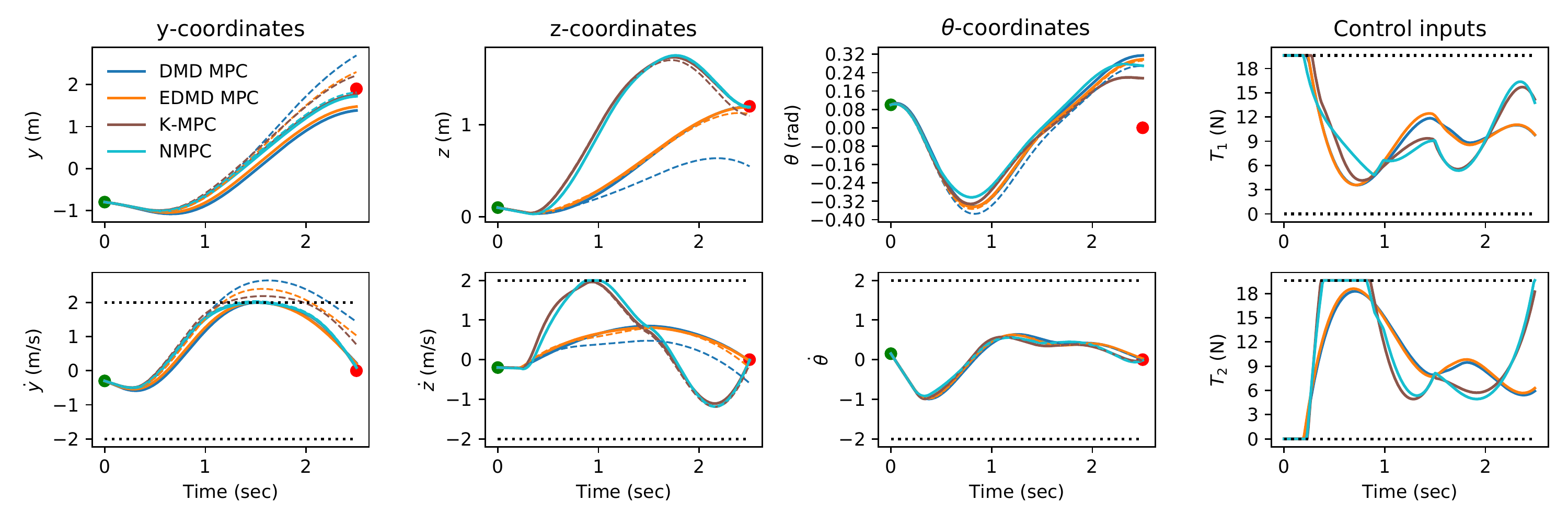}
    \caption{Trajectories generated with MPCs based on DMD, EDMD, and bEDMD models. True model-based NMPC used as benchmark. (Black dotted lines - state/actuation constraints, dashed lines - open loop simulation of generated trajectories).}
    \label{fig:trajectory_gen}
    \vskip -0.1 true in
\end{figure*}

As discussed in Section \ref{sec:sqp}, the SQP algorithm requires an initial guess of the solution $\mb x_i^{\text{init}}, \mb u_i^{\text{init}}$. Selecting a initial guess that is sufficiently close to the true optimal solution is essential for the algorithm to converge fast and reliably \cite{Gros2020}. It is well known that the receding horizon nature of MPC can be exploited to obtain excellent initial guesses. At a time instant $i$, this can be achieved by \textit{shifting} the NMPC solution from the previous timestep $i-1$ and by updating the guess of the final control input. Under certain conditions, a locally stable controller enforcing state and actuation constraints can be designed allowing feasibility of the initial guess to be guaranteed \cite{Rawlings2012}. Typically, simpler approaches are taken such as simply adding a copy of the final control signal and calculating the implied final state using the dynamics model
\begin{align} \label{eq:shift_procedure}
    \begin{split}
        &\mb u_{i,k}^{\text{init}} = \mb u_{i-1,k+1}, \quad k = 0, \dots, N-2,\\
        &\mb x_{i,k}^{\text{init}} = \mb x_{i-1,k+1}, \quad k = 0, \dots, N-1,\\
        &\mb u_{i,N-1}^{\text{init}} = \mb u_{i,N-2}^{\text{init}},
        \quad \mb x_{i,N}^{\text{init}} = f_d(\mb x_{i,N-1}^{\text{init}}, \mb u_{i,N-1}^{\text{init}}).
    \end{split}
\end{align}
If the previous solution $\mb x_{i-1}, \mb u_{i-1}$ is feasible, the shifted solution will also be feasible for all but the last timestep. 

\subsection{Calculating the linearized system matrices}
As a result of the bilinear structure of the dynamics model, the linearization can be efficiently computed for a given initial guess. The linearization at each timestep $k=0,\dots,N-1$ of the initial guess is obtained by directly calculating the partial derivatives as described in (\ref{eq:linearization})
\begin{align} \label{eq:koopman_linearization}
    \begin{split}
        A_{i,k} &= F + \sum_{j=1}^m G_j (\mb u_{i,k}^{\text{init}})^{(j)}, \,\, B_{i,k} = \big [ G_1 \mb z_{i,k}^{\text{init}} \dots G_m \mb z_{i,k}^{\text{init}} \big ]\\
        r_{i,k} &= F\mb z_{i,k}^{\text{init}} + \sum_{j=1}^m G_j\mb z_{i,k}^{\text{init}} (\mb u_{i,k}^{\text{init}})^{(j)} - \mb z_{i,k+1}^{\text{init}}.
    \end{split}
\end{align}
Consequently, the linearized dynamics matrices can be obtained by simple matrix multiplication and addition with the dynamics matrices of the Koopman model and the matrices containing the initial guesses of $\mb z_i^{\text{init}}, \mb u_i^{\text{init}}$.  

\section{Simulated Quadrotor Learning and Control}
\label{sec:result}

\subsection{System and data collection details}

Consider a planar quadrotor with states $\mathbf{x} = [y \, z \, \theta \, \dot{y} \, \dot{z} \, \dot{\theta}]^T$,
\begin{equation} \label{eq:quad_dynamics}
    \begin{bmatrix} \ddot{y} \\ \ddot{z} \\ \ddot{\theta} \end{bmatrix} 
    = \begin{bmatrix}
    0\\-g\\0
    \end{bmatrix} + 
    \begin{bmatrix}
    -(1/m)\text{sin}\theta & -(1/m)\text{sin}\theta\\
    (1/m)\text{cos}\theta & (1/m)\text{cos}\theta\\
    -l_{arm}/I_{xx} & l_{arm}/I_{xx}
    \end{bmatrix} 
    \begin{bmatrix}
    T_1 \\ T_2
    \end{bmatrix} ,
\end{equation}
where $y,z$ describe the horizontal and vertical position in a fixed reference frame, $\theta$ is the orientation, $T_1, T_2$ are the propeller thrusts, $g$ is the acceleration of gravity, $m$ is the vehicle mass, $l_{arm}$ is the distance from the vehicle's center of mass to the propeller axis, and $I_{xx}$ is the rotational inertia. 

To collect data, a nominal LQR controller is designed for the linearized dynamics at hover. Since the system is underactuated, learning trajectories are generated from a MPC based on the linearized dynamics. However, any controller can be used and the method does not need a known model linearization. Additionally, exploratory Gaussian white noise is added to aid sufficient excitation. The learning data set is collected as follows. First, an initial condition $ x_0$ and final condition $ x_f$, are sampled uniformly at random from the interval $y,z \in [-2, 2]^2, \theta \in [-\pi/3, \pi/3], \dot{y}, \dot{z},\dot{\theta} \in [-1,1]^3$. Then, 2-second long trajectories link $ x_0$ to $ x_f$, which are tracked via the LQR-controller. This process is repeated 100 times as state and actuation data is captured at 100 hz. 

To compare our method against the state-of-the art of Koopman-based learning methods, we trained three separate models, dynamic mode decomposition (DMD) \cite{Tu2014}, extended DMD (EDMD) \cite{Li2017}, and the method of Section \ref{sec:bedmd}, denoted bilinear EDMD (bEDMD). Assuming that the input forces enter through rotation matrices, we chose a simple dictionary of 27 functions for both the EDMD and bEDMD consisting of the state vector and monomials of the $\theta, \dot{\theta}$ up to the third order multiplied by $1, \cos(\theta), \sin(\theta)$, $\phi( x) = [1, \, y, \, z, \, \theta, \, \dot{y}, \, \dot{z}, \, \dot{\theta}, \, \theta
^2, \, \theta, \, \dot{\theta},\, \dots, \, \dot{\theta}^3, \cos\theta]^T$. $l1$-regularization tuned with cross-validation was also applied to each method. Code for learning and control is implemented in Python (\texttt{github.com/Cafolkes/koopman-learning-and\\-control}) and the dynamics are simulated using 5th order Runge-Kutta in \texttt{scipy}.


\subsection{Open loop prediction}

\begin{table}[t]
    \centering
    \begin{tabular}{p{4cm}|ccc}
        \hline
         & DMD & EDMD & bEDMD\\
         \hline
         Mean squared error & 8.71e-2& 5.60e-2 & 7.53e-3 \\
         Improvement vs DMD &  & 35.75 \% & 91.35 \%\\
         Improvement vs EDMD & & & 86.54 \%\\
         \hline
         Standard deviation & 2.79e-1 & 2.36e-1 & 8.66e-2 \\
         Improvement vs DMD & & 15.27 \% & 68,94 \% \\
         Improvement vs EDMD &&& 63.35 \%\\
         \hline
    \end{tabular}
    \caption{Prediction error of DMD, EDMD, and bEDMD models.}
    \label{tab:open_loop}
\end{table}

We first evaluate our method's prediction performance. A test data set is generated the same way as the training set. Then, the control sequence of each test trajectory is simulated forward with each of the learned models. The mean and standard deviation of the error between the true and predicted evolution over the trajectories are reported in Table \ref{tab:open_loop}. The experimental results support the theory: the mean and standard deviation of the error is reduced by 86-91 percent and 63 to 69 percent, respectively, compared to DMD and EDMD. bEDMD better captures the nonlinearities in the actuation matrix that drives the $(y,z)$-dynamics. 

\subsection{Trajectory generation and closed loop control}
\begin{figure*}[t]
    \centering
    \subfloat{\includegraphics[width=\textwidth]{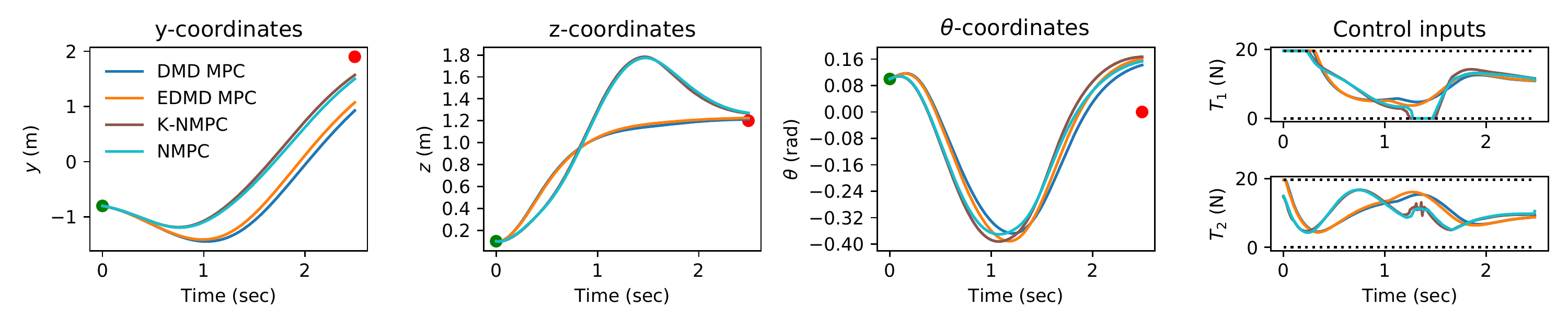}}\\
    \vspace{-0.5cm}
    \subfloat{\includegraphics[width=\textwidth]{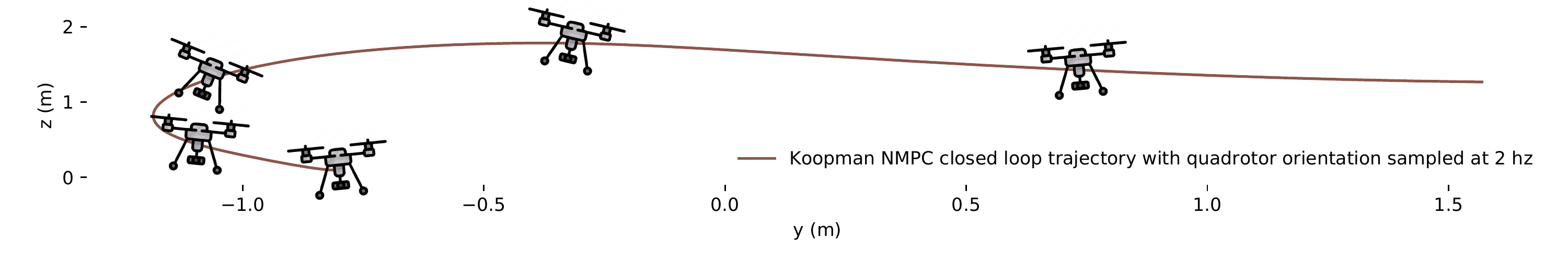}}
    \caption{Closed loop control with MPCs based on DMD, EDMD, and bEDMD models. True model-based NMPC used as benchmark.}
    \vskip -0.1 true in
    \label{fig:closed_loop}
\end{figure*}

To study trajectory generation and control, we first design MPCs for each of the learned models. For the linear (DMD) and lifted linear (EDMD) models, a linear MPC is designed. Then, the Koopman NMPC (K-NMPC) is designed as described in Section \ref{sec:mpc}. Finally, as a benchmark we implement NMPC using the true dynamics (\ref{eq:quad_dynamics}) based on Section \ref{sec:nmpc}. Each controller is based on a discrete-time model with sampling length 10 ms. All the optimization problems are solved with \texttt{OSQP} \cite{Stellato2018}. We initially study the ability of each  controller to generate high quality trajectories. One hundred trajectories (2.5 second duration) are designed to move the system from $x_0$ to $x_f$, sampled uniformly at random from the interval $y,z \in [-2, 2]^2, \theta \in [-0.1, 0.1], \dot{y}, \dot{z},\dot{\theta} \in [-1,1]^3$. The Frobenius norm of the control inputs over the prediction horizon is minimized and a terminal state constraint is added to each of the controllers to require that the desired position is reached. Finally, the velocities are constrained to have magnitude less than 2, $\dot{y}, \dot{z}, \dot{\theta} \in [-2,2]$, and the thrust of each propeller is limited, $T_1,T_2 \in [0, 2T_{hover}]$.

The generated trajectories from one of the experiments (solid lines) along with the open-loop simulation of the true dynamics with the control sequence of each designed trajectory (dashed lines) are depicted in Figure \ref{fig:trajectory_gen}. Table \ref{tab:trajectory_gen} presents summary statistics from 100 experiments: total control effort (as measured by the Frobenious norm and normalized by the NMPC control effort), the terminal state error (the Euclidean distance betwen the final open loop stimulation state and the desired state), and the number of SQP iterations needed by K-NPC and NMPC.  The open loop simulation reveals that the trajectories resulting from the DMD and EDMD models are not realizable, leading to significant mean terminal state errors of 2.24 and 2.46, respectively. K-NMPC has a significantly lower mean error of 0.70, and, more importantly, captures the idealized behavior of NMPC, even though it is completely data-driven.

\begin{table}[b]
    \centering
    \begin{tabular}{p{2cm}|C{0.9cm}C{0.9cm}C{1.3cm}C{1.3cm}}
        \hline
         & DMD & EDMD & bEDMD & Benchmark \\ 
         & (MPC) & (MPC) & (K-NMPC) & (NMPC)\\
     \end{tabular}
     \begin{tabular}{p{2cm}|C{0.28cm}C{0.28cm}C{0.28cm}C{0.28cm}C{0.4cm}C{0.4cm}C{0.4cm}C{0.4cm}}
        & mean & std & mean & std & mean & std & mean & std\\
         \hline
         Control effort & 0.97 & 0.02 & 0.97 & 0.06 & 1.01 & 0.01 & 1.00 & 0.00 \\
         Terminal error & 2.24 & 1.50 & 2.46 & 1.77 & 0.70 & 0.31 & 0.36 & 0.17\\
         SQP iterations &  &  & & & 21.88 & 11.57 & 9.33 & 9.63\\
         \hline
     \end{tabular}
    \caption{Summary statistics over 100 experiments of the MPC trajectory cost, error, and SQP iterations.}
    \label{tab:trajectory_gen}
\end{table}

Finally, we study closed loop control behavior  of each control approach over the same 100 initial and terminal conditions. Each of the MPCs use a 0.5 second prediction horizon with sampling length 10 ms and a quadratic state penalty and control input cost. The Koopman NMPC and NMPC controllers are initialized by solving each of the NLPs to convergence with the SQP algorithm before only a single SQP iteration is performed at each timestep in closed loop. 

The traces resulting from each controller are presented in Figure \ref{fig:closed_loop} for one of the experiments. Furthermore, summary statistics over the 100 experiments of the realized cost (as measured by the total trajectory cost, normalized by the NMPC cost), and the computation time at each timestep of each of the controllers are reported in Table \ref{tab:closed_loop}. Because closed loop operation can correct for model errors, the performance difference between the controllers is smaller than for the trajectory generation case. The controllers based on the DMD and EDMD models achieve a 4 and 2 percent higher cost than the NMPC, respectively. The K-NMPC again closely follows the behaviour of the NMPC.

The linear and lifted linear MPCs require less computational effort than the SQP-based approaches with an average computation time of 2 and 7 ms, respectively. K-NMPC requires somewhat higher computational effort than NMPC for this system with an average of 13 ms compared to 7 ms. This is dominated by longer solution time of the QP because a higher number of variables and constraints as a result of the lifting. The relative computational effort between K-NMPC and NMPC will ultimately depend on the complexity of linearizing the nonlinear model for NMPC, which can be expensive for complicated models, and the lifting dimension of the Koopman bilinear model. Finally, we note that even with a relatively simple python implementation the controllers are approaching real-time capability.

\begin{table}[b]
    \centering
    \begin{tabular}{p{2cm}|C{0.9cm}C{0.9cm}C{1.3cm}C{1.3cm}}
        \hline
         & DMD & EDMD & bEDMD & Benchmark \\ 
         & (MPC) & (MPC) & (K-NMPC) & (NMPC)\\
     \end{tabular}
     \begin{tabular}{p{2cm}|C{0.28cm}C{0.28cm}C{0.28cm}C{0.28cm}C{0.4cm}C{0.4cm}C{0.4cm}C{0.4cm}}
        & mean & std & mean & std & mean & std & mean & std\\
         \hline
         Realized cost & 1.04 & 0.05 & 1.02 & 0.03 & 1.00 & 0.00 & 1.00 & 0.00 \\
         Comp time (ms) & 1.50 & 0.34 & 7.40 & 1.78 & 13.14 & 8.35 & 6.99 & 1.11\\
         \hline
     \end{tabular}
    \caption{Summary statistics over 100 experiments of the MPC closed loop control cost and computation times.}
    \label{tab:closed_loop}
\end{table}


\section{Conclusion}
\label{sec:conclusion}

This paper presented a method to combine the learning of lifted bilinear models based on Koopman spectral theory with nonlinear model predictive control. This combination enables flexible data-driven control design with the many advantages of NMPC, such as state and actuation constraint satisfaction and optimality with respect to intuitive performance specifications. Through a simulated planar quadrotor example, we demonstrated the advantages of learning lifted bilinear models over lifted linear models to capture control-affine dynamics.  We also showed that the resulting data-driven controller could  achieve similar performance to the case of NMPC design with an exactly known dynamic model. 

Future work includes data-driven learning of the dictionary functions in the lifted bilinear model setting and applying the method experimentally to a multirotor drone. Learned dictionary functions will enable similar or better performance with more compact models (i.e. fewer functions in the dictionary). As a result, higher-dimensional system dynamics can learned with the method while maintaining acceptable computation times for the K-NMPC.


\bibliography{references_fixed} 
\bibliographystyle{ieeetr}
\end{document}